\documentclass[conference]{IEEEtran}
\IEEEoverridecommandlockouts
\usepackage{cite}
\usepackage{float}
\usepackage{todonotes}
\usepackage{amsmath,amssymb,amsfonts}
\usepackage{algorithmic}
\usepackage{graphicx}
\usepackage{subfigure}
\usepackage{textcomp}
\usepackage{xcolor}
\usepackage{booktabs,chemformula}
\def\BibTeX{{\rm B\kern-.05em{\sc i\kern-.025em b}\kern-.08em
    T\kern-.1667em\lower.7ex\hbox{E}\kern-.125emX}}

\begin{document}
%
\title{
A Reinforcement Learning Environment for Multi-Service UAV-enabled Wireless Systems
}

\author{\IEEEauthorblockN{Damiano Brunori}
\IEEEauthorblockA{Sapienza Univ. of Rome\\
brunori@diag.uniroma1.it}
\and
\IEEEauthorblockN{Stefania Colonnese and Francesca Cuomo}
\IEEEauthorblockA{Sapienza Univ. of Rome\\
DIET Dept.\\
stefania.colonnese@uniroma1.it\\francesca.cuomo@uniroma1.it}
\and
\IEEEauthorblockN{Luca Iocchi}
\IEEEauthorblockA{Sapienza Univ. of Rome\\
DIAG Dept.\\
iocchi@diag.uniroma1.it}}


%


\maketitle

\begin{abstract}
We design a multi-purpose environment for autonomous UAVs offering different communication services in a variety of application contexts (e.g., wireless mobile connectivity services, edge computing, data gathering).
We develop the environment, based on OpenAI Gym framework, in order to simulate different characteristics of real operational environments and we adopt the Reinforcement Learning to generate policies that maximize some desired performance.
The quality of the resulting policies are compared with a simple baseline to evaluate the system and derive guidelines to adopt this technique in different use cases.
The main contribution of this paper is a flexible and extensible OpenAI Gym environment, which allows to generate, evaluate, and compare policies for autonomous multi-drone systems in multi-service applications. This environment allows for comparative evaluation and benchmarking of different approaches in a variety of application contexts.
\end{abstract}

\begin{IEEEkeywords}
UAV networks, multi-service UAV, reinforcement learning.
\end{IEEEkeywords}

%
\IEEEpeerreviewmaketitle

\section{Introduction}
In the next years, a boost of advanced mobile services is expected, by the integration of a different variety of devices that may exploit the access network not only for throughput based services but also for support in computing tasks. A key role will be played by Unmanned Aerial Vehicles (UAV), as envisaged by 3GPP for 5G and beyond networks \cite{3gpp22}, as we expect UAVs to provide not only high throughput but also massive data processing. In this case, it is beneficial  to exploit either  network-based processing  or mobile computing. Furthermore, UAVs are expected to be exploited  to provide  messaging/sensing services to connected devices (things) \cite{uav_ioe}. Thereby, we address in this paper, in a holistic way, the provisioning of three kinds of services through UAVs: i) throughput provisioning, ii) edge computing) and iii) data collection.

Specifically, we present a framework based on Reinforcement Learning (RL) to make UAVs (drones in our case) learn how to provide these services by maximizing some specified Quality of Experience (QoE) of the users, while minimizing resources (e.g., battery energy) at the same time\footnote{https://damianobrunori.github.io/}. 

To optimize the desired performance, UAVs have to choose proper actions to be executed during their missions. In this context, the difficulty of providing exact models of the environment makes RL techniques very useful, as agents can learn from past experience, rather than computing solutions from a given model.
RL approaches however require the execution of many trials that cannot be done with real UAVs, therefore environments simulating relevant UAV behaviours are needed in order to train the policies that will then be run on real UAVs. 

Defining a multi-UAV environment for providing multiple services in an environment can be a complex task depending on the required level of reality. Moreover, the environment should be easy to use and to extend to match the different needs of researchers studying similar use cases.
Finally, when used as a system for RL, the environment should provide the necessary features to enable the execution of RL algorithms. Generating mission specific behaviors, like the ones presented in this paper, is also a key goal of the EU SESAR project BUBBLES\footnote{http://bubbles-project.eu/} on which this work is inspired.

The contribution of this paper is thus to describe the development of a flexible and extendable Multi-UAV Multi-Service environment compliant with the \textit{OpenAI Gym} environment, allowing the setting of different operating use-cases and enabling to run different RL algorithms. The environment is designed with specific attention to scalability and extendability, in order to be useful to other researchers for new experimental use-cases. Moreover, compliance with \textit{OpenAI Gym} allows exploiting many resources developed in such an environment and to create common benchmarks that will enable comparative analysis of different techniques and results.

\section{Related Work}
Many recent papers are focused on the use of UAVs for providing a variety of services and using RL techniques. In the following, we summarize some papers strictly related to the issues covered by the proposed environment.

\subsection{Multiple UAVs and Resource Allocation}
Use cases and inter-operations with the cellular networks are analyzed in 3GPP technical specification \cite{3gpp22}, with particular reference to possible applications such as the adoption of UAV to support high resolution video live broadcast application, e.g. with 360 degree spherical view camera system carried on a UAV, as well as the adoption of UAVs swarms in logistic. In \cite{mishra20} the authors describe how UAVs swarms can be integrated in 5G and beyond cellular systems either as relays, BSs or users. In \cite{jiang19} cache-enabled UAV is used to assist mobile-edge computing and the best 3D position among IoTs is searched to maximize data throughput. As for mobile streaming service, UAVs can provide flexible resource allocation; the problem of maximizing the QoE (minimize freeze time) in the mobile streaming services provided through UAVs is solved in \cite{chentvt20}, where the problem is addressed in terms of power/bandwidth allocation. Referring to UAVs swarm, they can also provide a service of wireless power transmission for multiple energy receiver; this service is described in \cite{wang17}, \cite{zhang19}, and \cite{xu18}. In the latter, a trajectory is designed to maximize the amount of energy transferred to all ERs during a finite charging period. In \cite{wang17},\cite{zhang19}, alongside the power transfer service, UAVs also provide clients (IoTs) edge-computing, with a binary \cite{wang17} or a partial \cite{zhang19} tasks offloading.

\subsection{Path Planning and RL Approaches}
Optimal joint resource allocation and path planning are definitively relevant for UAV swarms development, and it may leverage learning for fast and flexible adaptation to the environment as discussed in \cite{zheng20}. Path planning forms the basis of any resource allocation algorithms for multi-UAV systems, and it poses specific challenges, as detailed in \cite{aggar20}. One possible application for which path planning is needed is that of energy-efficient mobile computing, as in \cite{zhou18}. Other works (as \cite{zhang20}) focus on data gathering taking into account also the energy efficiency, where deep-learning techniques are used: here is planned the best path for both a UAV and a mobile charging station. \cite{zhou19} uses Deep Learning to pre-train an UAV employed to collect data from various IoTs devices trying to minimize to Age of Information to enhance data freshness. Among others, machine learning techniques are the most relevant for optimizing resource allocated to multiple UAVs in providing communications services; a survey dealing with these issues can be found in \cite{sensorml19}.

\subsection{Discussion}
Although many approaches have been proposed to address the deployment of multiple UAVs for delivering services, all these methods are based on specific settings (using specific simulators, environments, and experimental procedures) that are difficult to replicate and to be compared one each other.

The \textit{Multi-UAV Multi-Service OpenAI Gym} environment proposed in this paper has the goal of unifying these efforts in a common setting for developing and comparing multi-UAV RL techniques in different use-cases.

Moreover, in contrast with other Open AI gym environments modelling UAVs (such as gym-pybullet\footnote{https://github.com/JacopoPan/gym-pybullet-drones}), the \textit{Multi-UAV Multi-Service OpenAI Gym} presented here is focused on high-level interactions with the environment including modelling of users receiving services from the UAVs.

The presented environment allows us to easily model a Multi-UAV multi-service problem to find a behaviour of the UAV team according to some specific QoE performance metrics.

The problem discussed in this paper is different with respect to the literature mainly due to the UAV motion. Many other papers refer to UAV motion as an high-level motion, meaning that they are able to move from a cluster centroid (of users) to another one, assuming that UAVs know exactly where the users are located. This assumption is removed in the problem considered in this paper, thus UAVs can perform smoother movements from a cell (of the considered area) to another one, making the scenario more realistic. This kind of UAV motion, combined with the usage of our \textit{OpenAI Gym} environment, allows for easy reproducibility of the results, for comparison with other methods and can be extended to other different multi-service use-cases.

\section{Multi-UAV Multi-service environment}
The developed Multi-UAV Multi-service framework supports a variety of settings.
The environment involves a volume of operations which is partitioned into $K$ small cells whose size $A_k,k=0,\cdots K-1$ can be flexibly set according to the desired resolution per cell\footnote{A continuous visual representation of the environment is also implemented, but it is not discussed in this paper.}.
Obstacles can be randomly generated by setting the coverage percentage of the obstacles for the whole operational volume. Maximum building height can be selected and, according to this choice, different heights for each building are randomly generated. 

We can set the presence of $C$ clusters of users, whose centroid locations are generated by drawing samples from a uniform distribution; we can also set $U$ users, spread out along the generated centroids by using a truncated (bounded) normal distribution.
Each user requests one out of $N_{serv}$ services (as throughput request, edge-computing or data gathering), with some parameters stored as attributes of each user. 

In the environment considered in this paper, $N$ drones can be deployed, and $M$ Charging Stations (CS) are made available at locations placed in points made equidistant from the middle of the operational volume. These charging stations are placed at equal distance one from another. Drones are supposed to serve users while flying (except for some particular cases explained later on) and their performance are evaluated according to predefined metrics. 

Each drone is associated with an \textit{agent} characterized by a state representation (including its position and other desirable state variables) and an action space (that can vary according to the operational scenario).

The following \textit{general assumptions} are considered:
\begin{itemize}
        \item Multi-agent distributed system: UAVs do not know anything about other agents and do not communicate with each other;
        \item UAVs do not have a detailed map of the scenario, thus they implicitly learn to avoid obstacles;
        \item UAVs have a given footprint (i.e., a circular range around them) and each UAV can detect and serve only users inside it;
        \item in the 2D setting, the UAVs are assumed to fly at a unique altitude above the tallest building, thus without obstacles.
\end{itemize}

The following \emph{configurations} can be set:
\begin{itemize}
    \item static or dynamic service requests;
    \item stationary/moving users;
    \item perfect/noisy estimate of UAVs position;
    \item single/multiple services;
    \item 2D or 3D setting;
    \item limited/unlimited UAV battery;
    \item limited/unlimited UAV bandwidth.
\end{itemize}

Other features can be easily introduced in our environment.
%
%

\section{Experimental Scenario}
As already mentioned, the developed
Multi-UAV Multi-service environment can be configured to describe different operational use-cases (see Table \ref{table:cases_table}). In this section, we describe some configurations for which we report the relevant experimental analysis.

For these experiments, four UAV state space representations have been considered, with coordinates 
$(x_{UAV}, y_{UAV}, z_{UAV})$  and battery level $B_l$:

\begin{enumerate}
    \item $[(x_{UAV}, y_{UAV})]$ for 2D env. $\wedge$ NO battery lim.;
    \item $[(x_{UAV}, y_{UAV}), B_l]$ for 2D env. $\wedge$ battery lim.;
    \item  $[(x_{UAV}, y_{UAV}, z_{UAV})]$ for 3D env. $\wedge$ NO battery lim.;
    \item $[(x_{UAV}, y_{UAV}, z_{UAV}), B_l]$ for 3D env. $\wedge$ battery lim.
\end{enumerate}

The action space for each agents is:
\begin{itemize}
    \item \textit{flight} actions:
         forward, backward,
         right, left,
         hover,
         up, down (in 3D env.);
    \item \textit{charging} actions (only when considering limited battery):
         goto\_charging\_station,
        charge.
\end{itemize}

\begin{table}[h]
\footnotesize
    \caption{Adopted scenario parameters.}
    \centering
    \begin{tabular}{|l|l|ll|}
        \hline\hline
        \multicolumn{1}{c|}{Parameters} & \multicolumn{1}{c}{Values} \\
        \hline\hline
        \multicolumn{1}{c|}{Max UAV speed} & \multicolumn{1}{c}{8.3 $m/s$}\\
        \multicolumn{1}{c|}{Max UAV acceleration} & \multicolumn{1}{c}{4 $m/s^2$}\\
        \multicolumn{1}{c|}{Number of cells} & \multicolumn{1}{c}{100 (10x10)}\\
        \multicolumn{1}{c|}{Cell dimension} & \multicolumn{1}{c}{240x240 m}\\
        \multicolumn{1}{c|}{Map dimension} &
        \multicolumn{1}{c}{2400x2400 m}\\
        \multicolumn{1}{c|}{Clusters radius} & \multicolumn{1}{c}{240-480 m}\\
        \multicolumn{1}{c|}{Battery autonomy} & \multicolumn{1}{c}{30 minutes}\\
        \multicolumn{1}{c|}{UAVs footprint radius} & \multicolumn{1}{c}{600 m}\\
        \multicolumn{1}{c|}{1 iteration} & \multicolumn{1}{c}{1 minute}\\
        \multicolumn{1}{c|}{1 epoch} & \multicolumn{1}{c}{30 minute}\\
        \hline
    \end{tabular}
    \vspace{0.1cm}
\label{tab:ScenarioFeatures}
\end{table}

Table \ref{tab:ScenarioFeatures} shows some parameters used to configure the use-cases evaluated in this paper.

For what concerns the battery consumption, it is assumed that each UAV loses 1 minute of battery autonomy for each action performed.
With a bang-coast-bang acceleration profile, each drone travels 240 meters in approximately 60 seconds, thus the average speed per motion actions is approximately 4 m/s.

\begin{table}[h]
\footnotesize
    \caption{Adopted notation.}
    \centering
    \begin{tabular}{l|l|ll}
        \hline\hline
        \multicolumn{1}{c|}{Parameters} & \multicolumn{1}{c}{Notation}\\
        \hline\hline
        \multicolumn{1}{c|}{Number of UAVs} & \multicolumn{1}{c}{$N$}\\
        \multicolumn{1}{c|}{Number of Users} & \multicolumn{1}{c}{$U$}\\
        \multicolumn{1}{c|}{Number of Covered Users} & \multicolumn{1}{c}{$U_{C_{i}}$}\\
        \multicolumn{1}{c|}{Battery Level of the i-th UAV} & \multicolumn{1}{c}{$B_{i}$}\\
        \multicolumn{1}{c|}{i-th Critical Battery level} & \multicolumn{1}{c}{$c_{i}$}\\
        \multicolumn{1}{c|}{Needed Battery for the i-th UAV to go to the CS} & \multicolumn{1}{c}{$n_{B_{i}}$}\\
        \multicolumn{1}{c|}{j-th Reward Function for the i-th UAV} & \multicolumn{1}{c}{$R^{i}_{j}$}\\
        \hline
    \end{tabular}
\label{tab:firstNotaion}
\end{table}

Notation for parameters used to describe the configuration of the scenario and the associated rewards are summarized in Table \ref{tab:firstNotaion}.

In these experiments, in addition to the framework general assumptions described in the previous section, we considered the following:
\begin{itemize}
    \item[i)] Motion:
            \begin{itemize}
                \item UAVs cannot fly over the tallest building;
                \item when a UAV goes toward a charging station, it follows an A* path planning algorithm (based on the charging stations and obstacles positions within the scenario);
                \item each agent timing law has a trapezoidal speed profile and a bang-coast-bang acceleration (as in \cite{col19}) when moving from a cell to another;
                \item each movement is considered from the center of a cell to the center of the next one.
            \end{itemize}
        \item[ii)] Charging:
            \begin{itemize}
                \item no services are provided when going to CS or charging;
                \item UAVs leave a CS only when the battery is fully recharged;
                \item 
                UAVs learn with RL when it is time to go back to a CS; then they will go back according to A* algorithm.
            \end{itemize}
\end{itemize}

\subsection{Reinforcement Learning configuration}
The implemented environment allows initializing the state-action matrix, i.e. the Q-table, as follows:

\begin{itemize}
    \item zero initialization;
    \item random initialization;
    \item maximum reward initialization;
    \item prior knowledge initialization.
\end{itemize}

For our training process, we used the random initialization approach.
We compared two different standard RL algorithms, Q-learning and SARSA algorithms, with $\epsilon$-greedy exploration strategy.

\begin{table*}[t]
\footnotesize
\centering
\caption{Features of analyzed cases. \textit{Dyn.} represents the service request, which can be dynamic or static; \textit{Multi} indicates if UAVs are providing multi-services or a single-service. \textit{bat.} indicates the availability of UAVs battery (infinite or limited), while \textit{Us. mot.} stands for \textit{Users motion}. The UAVs bandwidth \textit{BW} is expressed in MHz. Under \textit{N° of users}, if two numbers appear, the first one refers to the Q-learning application, while the second to the SARSA case; otherwise a single number refers to the application of both algorithms. \textit{T} and \textit{F} simply stand for \textit{True} and \textit{False}.}
  \begin{tabular}{cccccccccccc}
    \toprule
    Case & Env. dim & BW & bat. & N° of UAVs & N° of clust. & N° of us. & N° of CSs & Dyn. & Multi & Us. mot. & Pos. error \\
    \midrule
    1 & 2D & - & inf & 1 & 1 & 14 & - & F & F & F & F
        \\  \addlinespace[5pt]
    2 & 2D & - & inf & 2 & 2 & 19-20 & - & F & F & F & F
        \\  \addlinespace[5pt]
    3 & 2D & - & inf & 2 & 3 & 35-33 & - & F & F & F & F
        \\  \addlinespace[5pt]
    4 & 2D & - & lim & 2 & 3 & 3-24 & 2 & F & F & F & F
        \\  \addlinespace[5pt]
    5 & 3D & - & lim & 2 & 2 & 21-24 & 2 & F & F & F & F
        \\  \addlinespace[5pt]
    6 & 3D & 5 & lim & 2 & 2 & 21-27 & 2 & F & T & F & F
        \\  \addlinespace[5pt]
    7 & 3D & 5 & lim & 2 & 3 & 41-36 & 2 & F & T & F & F
        \\    \addlinespace[5pt]
    8 & 3D & 5 & lim & 3 & 4 & 48 & 3 & T & T & T & F
        \\  \addlinespace[5pt]
    9 & 3D & 5 & lim & 3 & 4 & 48 & 3 & T & T & T & T \\
    \bottomrule
  \end{tabular}
  \label{table:cases_table}
\end{table*}


The reward functions used in the experiments vary according to the specific use case, as different variables are taken into account.

Referring to the parameters described in Table \ref{tab:firstNotaion},
the reward function used for the use cases 1-3 is;

\begin{equation}
    \mathcal{R}_{1}^{i} = \frac{U_c}{\hat{U}} \text{, where } \hat{U} = \frac{U}{N}
\label{eq:reward1}
\end{equation}

The reward used for the use cases 4-5 is:

\begin{equation}
    \mathcal{R}_{2}^{i} = w_{s}r_u + w_{c}r_c 
\end{equation}

where:


\begin{equation}
w_{s} = 
        \begin{cases}
        1.0 \;\;\;
          \text{if}\ B_{i} > c_{1} \\
        0.8 \;\;\;
          \text{if}\ (c_{1} < B_{i} \leq c_{1}) \wedge (B_{i} > n_{B_{i}}) \\
        0.5 \;\;\;
          \text{if}\ (c_{3} < B_{i} \leq c_{2}) \wedge (B_{i} > n_{B_{i}}) \\
        0.2 \;\;\; 
          \text{if}\ (c_{4} < B_{i} \leq c_{3}) \wedge (B_{i} > n_{B_{i}}) \\
        0.0 \;\;\;  
          \text{otherwise, i.e. if}\ B_{i} \leq n_{B_{i}} 
        \end{cases}
\label{eq:cost_user_weights}
\end{equation}

\begin{equation}
w_{c} = 
        \begin{cases}
        0.0 \;\;\;
          \text{if}\ B_{i} > c_{1} \\
        0.2 \;\;\;
          \text{if}\ (c_{1} < B_{i} \leq c_{1}) \wedge (B_{i} > n_{B_{i}}) \\
        0.5 \;\;\;
          \text{if}\ (c_{3} < B_{i} \leq c_{2}) \wedge (B_{i} > n_{B_{i}}) \\
        0.8 \;\;\; 
          \text{if}\ (c_{4} < B_{i} \leq c_{3}) \wedge (B_{i} > n_{B_{i}}) \\
        1.0 \;\;\;  
          \text{otherwise, i.e. if}\ B_{i} \leq n_{B_{i}} 
        \end{cases}
\label{eq:cost_user_weightc}
\end{equation}

where $w_s$ and $w_c$ represent the weights parameters for the service and cost rewards,respectively; $r_{u}$ is computed as in (\ref{eq:reward1}) and the \text{reward cost} is:

\begin{equation}
    r_c = \frac{n_{B_{i}}}{B_{i}}
    \label{eq:cost_reward}
\end{equation}

\noindent 
Finally, the reward function used for the use cases 6-9 is:

\begin{equation}
    \begin{array}{l}
    \mathcal{R}_{3}^{i} = w_{s}\left(w_{u}r_u + w_{tr}s_{tr} + w_{ec}s_{ec} + w_{dg}s_{dg} \right) + \\ + w_{c}(r_{cp} + r_{cs})
    \end{array}
\label{eq:reward3}
\end{equation}

where $w_{s}$ and $w_{c}$ are computed as in (\ref{eq:cost_user_weights}), $r_{cp}$ is the same as (\ref{eq:cost_reward}) and $r_{cs}$ represents the cost due to the throughput request service (which is considered negligible for these use-cases, but that could be added in future); $w_{u}$, $w_{tr}$, $w_{ec}$, $w_{dg}$ are the weights corresponding respectively to the percentage of i) covered users (regardless of the service requested), ii) throughput requests covered, iii) edge computing requests covered, iv) data gathering requests covered.

\subsection{Service-oriented Performance metrics}
We chose three different assessment (or Quality of Experience-QoE) parameters:
\begin{itemize}
    \item $QoE_1$: percentage indicating how much a user is covered based on his/her time service request (i.e., “completion percentage” per service);
    \item $QoE_2$: indicates the iteration steps between the request and the complete provision of the service;
    \item $QoE_3$: the percentage of the users covered by the UAVs service.
\end{itemize}

Many others could be selected according to the desired goals.
Notice that during training, the agent moves only according to the reward (i.e., they do not follow $QoEs$ trends).  

\section{Experimental results}

\subsection{Parameter settings}

We studied and tested 9 different cases, by varying several parameter settings in order to progressively increase the complexity level of each scenario: environment dimension, number of UAVs and users, service request and users motion are just some of all the settings modified for each analyzed case. 
All the study cases are illustrated in Table \ref{table:cases_table}.
Let us remark that in these cases we have considered the UAVs as the only service access points. Still, the architecture can naturally include information about other service access points by suitable adapting the reward function.

\subsection{Performance analysis}

The performance was compared with a specific baseline behaviour. With \textit{baseline behaviour} here it is meant a \textit{static} behaviour which is always the same and that allows each drone to go to the closest charging station when its battery level is equal to $15\%$ of the overall capacity; the return is always managed through \textit{A*}. This baseline behaviour is based on the following steps:

\begin{enumerate}
    \item the first drone will follow a square trajectory on the map by moving two cells away from the edge of the considered area;
    \item the second one will move only along the $y$-axis of the considered area;
    \item the third one will move only along the $x$-axis of the considered area;
    \item and so on, with each next drone moving according to 1. and 2. starting from where it was initially placed. 
\end{enumerate}

In a 3D environment, the assumption that the drone cannot fly higher than the tallest building is removed. Thus in this case, in order to avoid a building, a drone will simply fly over it whatever the height of the obstacle is.

\begin{table}[h]
    \centering
    \begin{tabular}{|l|c|c|c|}
    \hline
    ~~ \textbf{Metric} ~~& \multicolumn{3}{|c|}{\textbf{Algorithm}} \\
       \hline
    &\textit{\textbf{Baseline}} & \textit{\textbf{Q-learning}} & \textit{\textbf{SARSA}} \\
    \hline
    \multicolumn{3}{|l|}{\qquad\quad\textit{\textbf{Case 1}}}\\
    \hline
    $QoE_1$ & 20 \(\%\) & 75 \(\%\) & 100 \(\%\) \\
    $QoE_2$ & 42 it. &  0 it. & 0 it. \\
    $QoE_3$ & 15 \(\%\) & 100 \(\%\) & 98 \(\%\) \\
    \hline
    \multicolumn{3}{|l|}{\qquad\quad\textit{\textbf{Case 2}}}\\
    \hline
    $QoE_1$ & 27 \(\%\) & 55 \(\%\) & 100 \(\%\) \\
    $QoE_2$ & 22 it. & 0 it. & 0 it. \\
    $QoE_3$ & 33 \(\%\) & 97 \(\%\) & 100 \(\%\) \\
    \hline
    \multicolumn{3}{|l|}{\qquad\quad\textit{\textbf{Case 3}}}\\
    \hline
    $QoE_1$ & 27 \(\%\) & 60 \(\%\) & 98 \(\%\) \\
    $QoE_2$ & 21 it. & 30 it. & 30 it. \\
    $QoE_3$ & 34 \(\%\) & 80 \(\%\) & 55 \(\%\) \\
    \hline
    \multicolumn{3}{|l|}{\qquad\quad\textit{\textbf{Case 4}}}\\
    \hline
    $QoE_1$ & 23 \(\%\) & 10 \(\%\) & 98 \(\%\) \\
    $QoE_2$ & 226 it. & 0 it. & 15 it. \\
    $QoE_3$ & 30 \(\%\) & 20 \(\%\) & 60 \(\%\) \\
    \hline
    \multicolumn{3}{|l|}{\qquad\quad\textit{\textbf{Case 5}}}\\
    \hline
    $QoE_1$ & 27 \(\%\) & 52 \(\%\) & 95 \(\%\) \\
    $QoE_2$ & 27 it. & 8 it. & 30 it. \\
    $QoE_3$ & 25 \(\%\) & 68 \(\%\) & 42 \(\%\) \\
    \hline
    \multicolumn{3}{|l|}{\qquad\quad\textit{\textbf{Case 6}}}\\
    \hline
    $QoE_1$ & 28 \(\%\) & 100 \(\%\) & 45 \(\%\) \\
    $QoE_2$ & 228 it. & 0 it. & 30 it. \\
    $QoE_3$ & 25 \(\%\) & 65 \(\%\) & 10 \(\%\) \\
    \hline
    \multicolumn{3}{|l|}{\qquad\quad\textit{\textbf{Case 7}}}\\
    \hline
    $QoE_1$ & 64 \(\%\) & 100 \(\%\) & 98 \(\%\) \\
    $QoE_2$ & 11215 it. & 20 it. & 100 it. \\
    $QoE_3$ & 13 \(\%\) & 65 \(\%\) & 50 \(\%\) \\
    \hline
    \multicolumn{3}{|l|}{\qquad\quad\textit{\textbf{Case 8}}}\\
    \hline
    $QoE_1$ & 68 \(\%\) & 100 \(\%\) & 98 \(\%\) \\
    $QoE_2$ & 10708 it. & 50 it. & 30 it. \\
    $QoE_3$ & 13 \(\%\) & 55 \(\%\) & 30 \(\%\) \\
    \hline
    \multicolumn{3}{|l|}{\qquad\quad\textit{\textbf{Case 9}}}\\
    \hline
    $QoE_1$ & 52 \(\%\) & 97 \(\%\) & 97 \(\%\) \\
    $QoE_2$ & 215184 it. & 25 it. & 35 it. \\
    $QoE_3$ & 9 \(\%\) & 45 \(\%\) & 25 \(\%\) \\
    \hline
    \end{tabular}
    \vspace{0.1cm}
    \caption{Values of the QOE metrics measured in the last training epoch in cases (1-9): comparison between SARSA and Q-learning algorithms w.r.t. the baseline behaviour.}
    \label{table:comparison_table}
\end{table}

Figure \ref{fig:TrainingComparison} shows the QoE metrics in two of the analyzed cases (Case 1 and Case 5), comparing Q-learning algorithms performance. Table \ref{table:comparison_table} provides instead a quick visual comparison of agents performance for each case of study. 

\renewcommand{\thefigure}{2}
\begin{figure*}[h]
  
  \begin{center}
%
        \subfigure{%
            \includegraphics[width=0.32\textwidth,height=0.16\textheight]{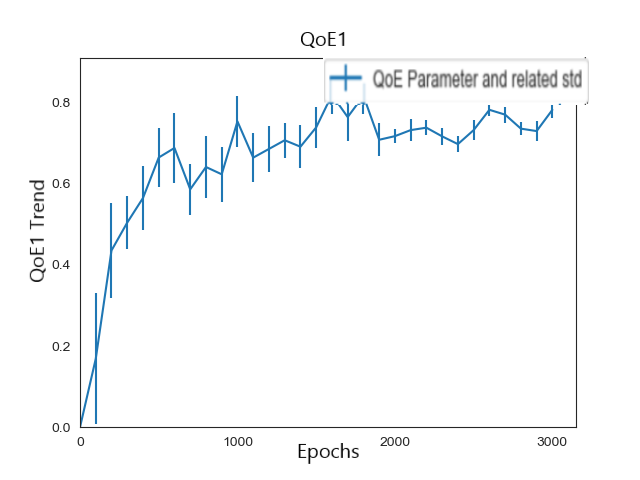}
        }%
       \hspace{-1.6\baselineskip}
        \subfigure{%
           \includegraphics[width=0.32\textwidth,height=0.16\textheight]{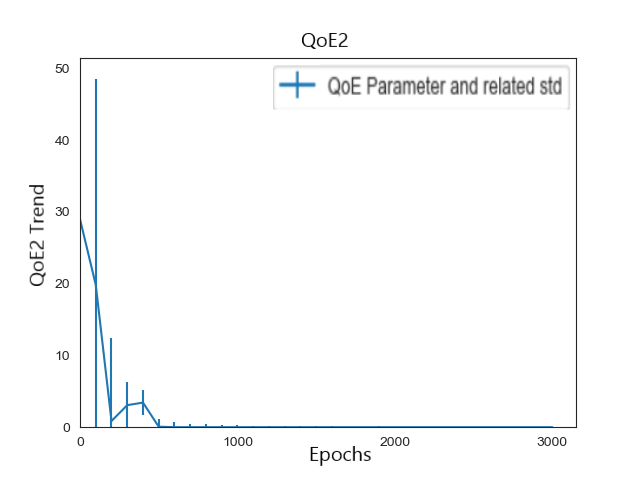}
        }
        \hspace{-1.6\baselineskip}
        \subfigure{%
           \includegraphics[width=0.32\textwidth,height=0.16\textheight]{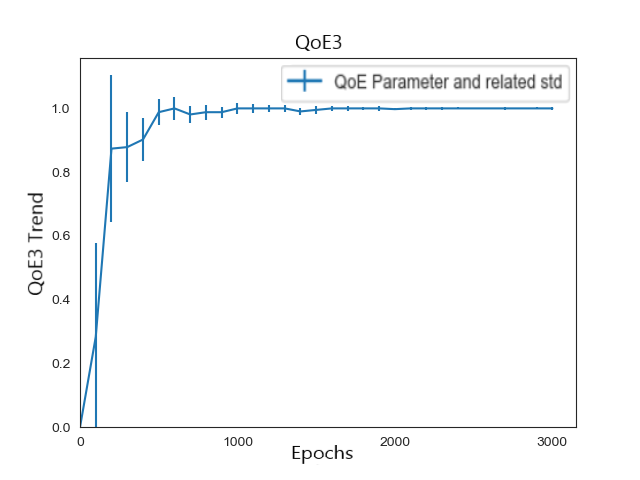}
        }
        \subfigure{%
            \includegraphics[width=0.32\textwidth,height=0.16\textheight]{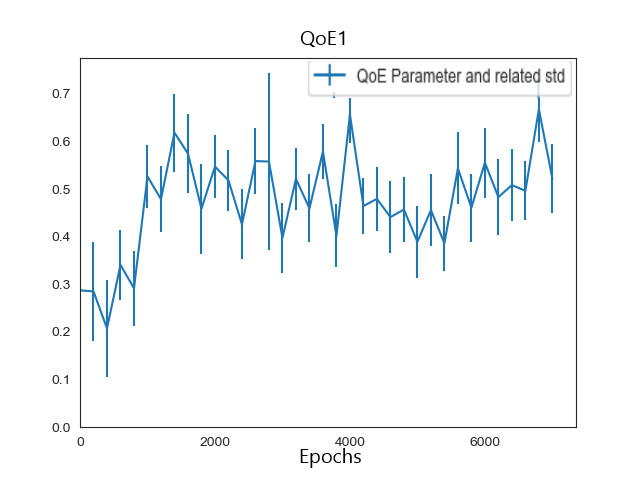}
        }%
        \hspace{-1.6\baselineskip}
        \subfigure{%
           \includegraphics[width=0.32\textwidth,height=0.16\textheight]{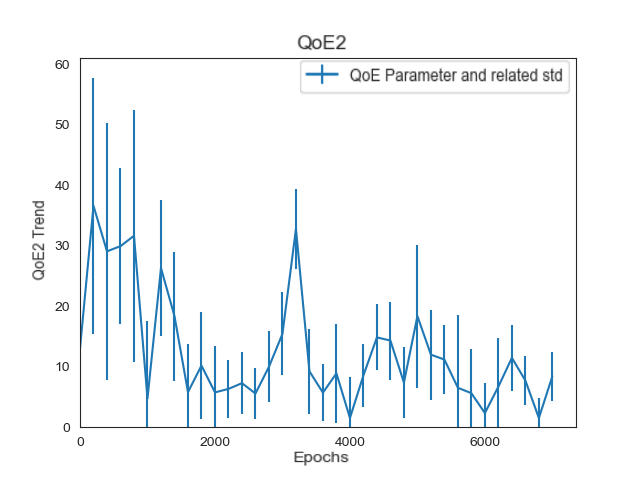}
        }
        \hspace{-1.6\baselineskip}
        \subfigure{%
           \includegraphics[width=0.32\textwidth,height=0.16\textheight]{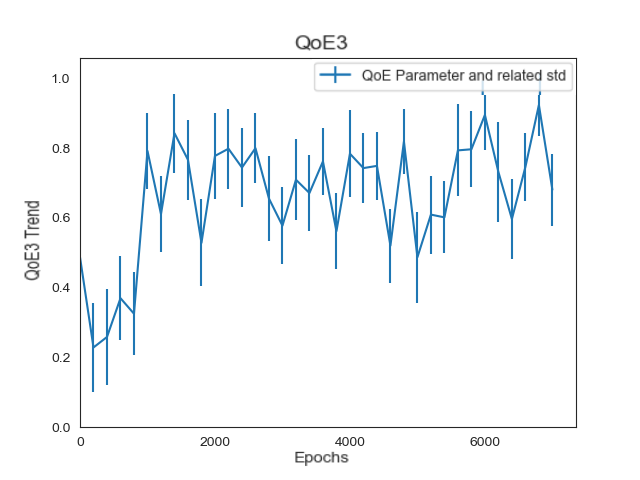}
        }
        \subfigure{%
           \includegraphics[width=0.38\textwidth,height=0.17\textheight]{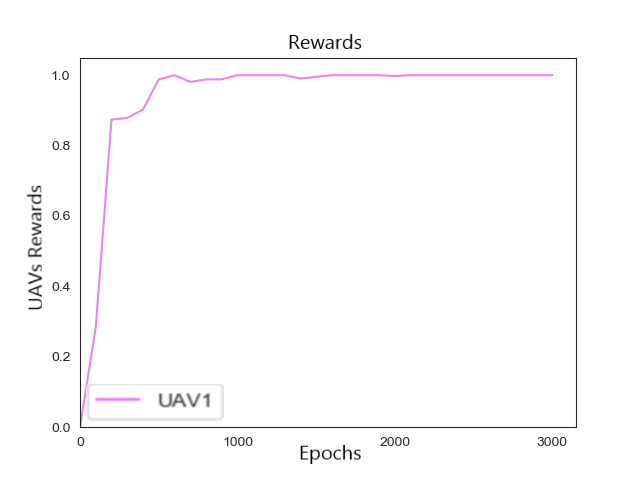}
        }
        \hspace{-1.6\baselineskip}
        \subfigure{%
           \includegraphics[width=0.38\textwidth,height=0.17\textheight]{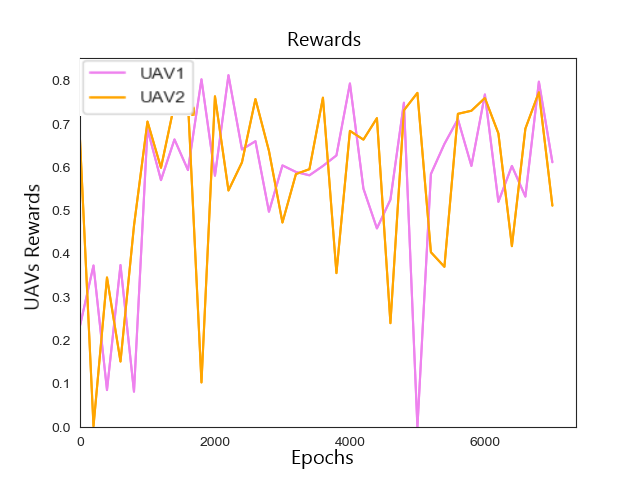}
        }
    \end{center}
    \caption{%
        QoEs (vertical lines represent the \textit{standard deviation}) and rewards trends when Q-learning algorithm is applied: the first horizontal row of graphs represents Case 1, the second one refers to Case 5, while the third one represents the rewards for Case 1 and Case 5 (from the left to the right). 
     }%
   \label{fig:TrainingComparison}
\end{figure*}

The performed experiments point out that the number of UAVs crashes dramatically increases when the error measure on the position of the drones is introduced. Moreover, the baseline behaviour is always the worst choice in terms of performance (but the best in term of safety) w.r.t. the RL approaches.


Our \textit{Multi-UAV Multi-Service OpenAI Gym} environment allows to configure and extend many different experiments and use-cases, by modelling problems similar to the one faced here. In this way, many different configuration tests and the related outcomes can be investigated and compared to improve the overall performance in each case of study. Figure \ref{fig:EnvironmentRepresentation} shows a screenshot of a training session related to a specific 3D scenario.

\renewcommand{\thefigure}{1}
\begin{figure}[H]
  \begin{center}
           \includegraphics[width=0.81\columnwidth]{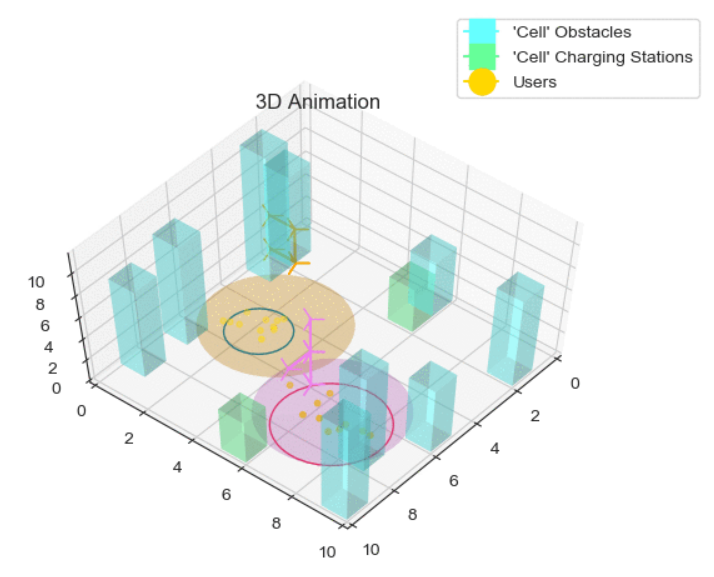}
    \end{center}
    \caption{%
        Snapshot of the a 3D environment. Colored crosses and lines represent respectively the agents and their current paths, filled circles are their footprints. 
     }%
   \label{fig:EnvironmentRepresentation}
\end{figure}

\section{Conclusion and Future Work}
In this work we have presented an environment to model Multi-UAV Multi-Service use-cases and to run comprehensive Reinforcement Learning algorithms. The architecture encompasses several scenario models and design choices. Experimental results show that  the architecture provides a flexible and scalable environment for the design of different service criteria. This paves the way for future works related to the multi-UAV application systems. Besides, the architecture modularity allows for extension in terms of: i) communication layer modelling (e.g. path loss models and fading in the considered environment), ii) adoption of neural network Deep RL algorithms for huge state space handling.

Finally, we believe that the \textit{Multi-UAV Multi-Service OpenAI Gym} environment can  boost reproducibility of results as well as comparison among different approaches in use-cases that are realistic in terms of modelling user-oriented services.

\section{Acknowledgements}
This paper has been partially supported by BUBBLES Project. BUBBLES 
project has received funding from the SESAR Joint Undertaking under 
the European Union’s Horizon 2020 research and innovation program 
under grant agreement No 893206. Research has been partially supported also by the ERC Advanced Grant WhiteMech (No. 834228) and by the EU ICT-48 2020 project TAILOR (No. 952215).

\end{document}